\documentclass[a4paper,12pt]{article}

\usepackage[truedimen,margin=20truemm]{geometry}

\usepackage{amsmath,amsthm,comment,amssymb}
\usepackage{graphicx}
\usepackage{color}
\newcommand{\Prob}{\mathrm{Prob}}

\newcommand{\Var}{\mathrm{Var}}

\newcommand{\cD}{\mathcal{D}}

\newcommand{\cF}{\mathcal{F}}

\newcommand{\cN}{\mathcal{N}}

\newcommand{\cU}{\mathcal{U}}
\newcommand{\cX}{\mathcal{X}}

\newcommand{\bR}{\mathbf{R}}

\newcommand{\argmin}{\mathop{\rm argmin}\limits}

\newcommand{\bV}{\mathbf{V}}
\newcommand{\rdo}{\mathrm{do}}
\newcommand{\indepe}{\mathop{\perp\!\!\! \!\perp}}
\newcommand{\bv}{\mathbf{v}}
\newcommand{\rFS}{\mathrm{FS}}
\newcommand{\rCF}{\mathrm{CF}}

\usepackage{algorithm}
\usepackage{algorithmic}
\usepackage{enumitem}

\usepackage[T1]{fontenc}
\usepackage[utf8]{inputenc}
\usepackage[font=small,labelfont=bf,tableposition=top]{caption}
\theoremstyle{plain}
\newtheorem{thm}{Theorem}

\newtheorem{prop}[thm]{Proposition}

\newtheorem{fact}[thm]{Fact}

\theoremstyle{definition}
\newtheorem{df}[thm]{Definition}

\newtheorem{assump}[thm]{Assumption}

\newtheorem{rmk}[thm]{Remark}

\usepackage[utf8]{inputenc} 
\usepackage[T1]{fontenc}    
\usepackage{hyperref}       
\usepackage{url}            
\usepackage{booktabs}       
\usepackage{amsfonts}       
\usepackage{nicefrac}       
\usepackage{microtype}      

\title{Causality and Robust Optimization}

\author{%
  Akihiro Yabe\\
  NEC corporation\\
  \texttt{a\_yabe@nec.com} 
}

\date{}

\begin{document}

\maketitle

\begin{abstract}
	A decision-maker must consider {\it cofounding bias} when attempting to apply machine learning prediction, and, while feature selection is widely recognized as important process in data-analysis, it could cause cofounding bias.
	A causal Bayesian network is a standard tool for describing causal relationships,
	and if relationships are known, then adjustment criteria
	can determine with which features cofounding bias disappears.
	A standard modification would thus utilize causal discovery algorithms for preventing cofounding bias in feature selection.
	Causal discovery algorithms, however, essentially rely on the faithfulness assumption,
	which turn out to be easily violated in practical feature selection settings.
	In this paper, we propose a meta-algorithm that can remedy existing feature selection algorithms in terms of cofounding bias.
	Our algorithm is induced from a novel adjustment criterion that requires
	rather than faithfulness, an assumption which can be induced from another well-known assumption of the causal sufficiency.
	We further prove that the features added through our modification convert cofounding bias into prediction variance.
	With the aid of existing robust optimization technologies that regularize risky strategies with high variance, then,
	we are able to successfully improve the throughput performance of decision-making optimization, as is shown in our experimental results.
\end{abstract}

\section{Introduction}
With recent advances in machine learning technology,
decision-making optimization aided by prediction 
has become ubiquitous in a variety of industries.
This paper considers decision-making conducted on the basis of batch learning and mathematical optimization, 
for which such a data-analysis pipeline is often called \emph{predictive optimization}~\cite{ito2018unbiased}.
Let us first introduce an example of the pipeline in price optimization~\cite{ito2016large}.
Let $x$, $y$, and $z$ denote, respectively, decision variables (product prices), target variables to be predicted (product demand), and external features (weather, temperature, etc.),
and suppose that one would like to maximize the revenue function $r(x,y)$
which is an inner product of price and demand vectors.
For this aim, given historical daily point-of-sales data $\cD = \{(x_d, y_d, z_d) \mid d=1,2,\dots,D \}$,
a typical learner would first apply a feature selection algorithm to compute subset $z_\kappa$ of external features $z$ for improving prediction performance, and next estimate sales demand prediction formula $\hat{y} = \hat{f}(x, z_{\kappa})$.
At the beginning of each day, then, the learner would input specific realization $\tilde{z}_{\kappa}$ of external features into a system, and it would optimize the pricing strategy $x$ that would maximize a revenue function $r(x,\hat{y})$ on the basis of $\tilde{z}_{\kappa}$ and prediction formula $\hat{y} = \hat{f}(x,\tilde{z}_{\kappa})$.
The general predictive optimization framework is applicable to a variety of applications,
such as portfolio optimization~\cite{qiu2015robust}, inventory optimization~\cite{bienstock2008computing}, and electricity auctions~\cite{kwon2012optimization}.

For decision-making optimization on the basis of a prediction formula,
one must be careful about \emph{cofounding bias},
which might make target variables unpredictable in terms of optimization.
Figure 1 shows a simple motivating example with three variables in predictive price optimization,
where feature selection causes cofounding bias.
A storekeeper is to decide the price of an umbrella
and weather is a cofounding variable affecting both price and demand; on rainy days, the demand for umbrellas is high,
and the storekeeper, knowing this, will raise prices accordingly.
Though increased prices would have a negative effect on demand, it would be less than the positive effect of rain.
Suppose we are given historical data and run a demand prediction algorithm.
A prediction model relating increased prices to increased demand would be simple, accurate, and, thus, preferable for a machine predictor.
If an optimizer increased price on the basis of this prediction model, however,
the demand might decrease unexpectedly, which is called cofounding bias.
The cofounding bias occurs since the rain node is deleted by the feature selector, is invisible to the optimizer, and thus behaves as a virtually unobserved cofounder.
This example demonstrates that 
a prediction model may not indicate the consequences of optimization under the existence of cofounding bias,
and this could be caused by a feature selection algorithm.

A causal Bayesian network~\cite{pearl2009causality,kalisch2014causal} is a well-known tool for describing causal relationships between variables,
and if such relationships are known,
then adjustment criteria~\cite{pearl1993bayesian,maathuis2015generalized,perkovic2015complete}
can inform that set of features with which one can avoid cofounding bias.
A significant amount of effort has thus been 
exerted on the study of causal discovery~\cite{spirtes2000causation,chickering2002learning,chickering2002optimal},
which aims to recover the structure of unknown networks from observational data.
In our predictive optimization setting,
the temporal context of the analysis pipeline indicates that
the set of direct causes of target variables satisfies the adjustment criteria,
and direct cause discovery algorithms~\cite{pena2007towards,gao2015local} 
are thus applicable.
In terms of feature selection,
a set of direct causes are known to maximize the performance of subsequent prediction~\cite{cawley2008causal,guyon2007causal}.
Direct cause discovery might thus be one of the most promising approaches for feature selection
in predictive optimization that simultaneously avoids cofounding bias
and improves prediction performance.

Existing causal discovery algorithms essentially rely on
a faithfulness assumption,
but that is easily violated in typical feature selection settings.
Faithfulness requires that every conditional independence can be read from the structure of a causal network, and this is often justified in the study of causal discovery since 
unfaithful parameterization has a measure zero and is thus unnatural.
For preprocessing of feature selection, however,
one often generates artificial features from original features via arithmetics
(quadratic features $z_i z_j$ from original features $z_i$ and $z_j$, for example,),
and this artificial generation violates the faithfulness condition.
Further,
in practice a causal discovery algorithm require "enough faithfulness"
for it to be verified with a given limited number of data.
The notion of $\lambda$-strong faithfulness has been studied under the normality assumption~\cite{kalisch2007estimating,uhler2013geometry,zhang2002strong},
and it characterizes the relationships among the amount of faithfulness, number of samples, and number of features.
These studies have shown that that the number of samples should be comparably larger than
the number of features,
while this might not the case in practical feature selection settings.
Thus, the faithfulness assumption cannot be justified in predictive optimization,
and, in fact,
causal discovery algorithms causes cofounding biases,
as is shown in our experiments.
\paragraph{Our contributions}
Our contributions are mainly two-fold. 
First, we present a novel adjustment criterion
that directly lead us to propose a meta-algorithm 
which can modify an existing feature selection algorithm so as 
to prevent cofounding bias, or to be admissible from the viewpoint of causality.
Our approach does not rely on the faithfulness assumption,
but instead relies on an assumption 
that the entire feature set $Z$ satisfies adjustment criteria,
which can be induced from the nonexistence of unobserved cofounders, called causal sufficiency assumption.
Intuitively speaking,
our approach can enjoy larger number of feature candidates that tend to include thorough cofounders,
while existing approaches relying on faithfulness
might suffer in a large number of candidates.
Our approach can thus be naturally applied to practical feature selection setting
dealing with larger number of features.

Secondly, we reveal the role of features additionally selected through the modification of our meta-algorithm in the predictive optimization pipeline.
Our meta-algorithm requires a feature selector
to adopt additional features which, though useless for improving prediction accuracy,
can reduce the cofounding bias.
Our theoretical analysis proves that,
under certain assumptions,
the sum of cofounding bias and variance is constant.
Thus the additional features can convert cofounding bias into prediction variance, which is measurable in practice and thus rather tractable.
With the aid of existing robust optimization technologies that regularize risky strategies with high variance, then,
we are able to successfully improve the throughput performance of a predictive optimization pipeline, as is shown in our experimental results.

Because of space limitation, all proofs are presented in the supplementary material.

\begin{figure}[t]
	\centering
	\includegraphics[width=0.8\linewidth]{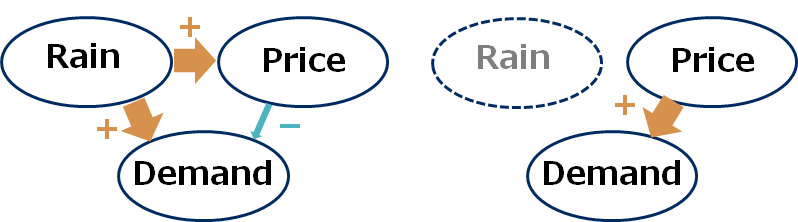}
	\caption{Example with three variables in price optimization. 
		The variables consist of a decision variable "price," a target variable "demand,"
		and an external feature "rain." The portion on the left shows the true causal model, 
		where price has a negative effect on demand, and rain has a positive effect on
		both price and demand.
		The portion on the right shows an accurate prediction model,
		in which high price is related to high demand, a so called "spurious correlation."
		The rain variable is invisible to the predictor and optimizer and,
		thus, behaves as an unobserved cofounder.
	} \label{fig1}
\end{figure}
\section{Predictive Optimization Problem}\label{sec_prob}
Our general predictive optimization problem,
consisting of feature selection, prediction,
and optimization,
is introduced in this section.
Let $x \in \cX \subseteq \bR^M$ be a vector of decision variables, where $\cX$ is an optimization domain.
Also, let $y \in \bR^N$ be a vector of target variables,
and $z \in \bR^K$ be a vector of external features.
Let $r : \bR^M \times \bR^N \to \bR$ be a given objective function.
The aim of predictive optimization here is to select optimum decision variable $x$ to minimize $r(x,y)$ 
on the basis of features $\tilde{z} \in \bR^n$ obtained in advance.
If exact prediction formula $y = f(x,z)$ is known,
then the problem can be formulated as a mere mathematical optimization problem:
\begin{align}
\min_{x \in \cX} r(x, y) \quad \text{s.t. } y=f(x,\tilde{z}). \label{idealOpt}
\end{align}
Since the exact prediction formula $y = f(x,z)$ is unavailable in practice,
we have to estimate it from historical data $\cD = (x^d,y^d,z^d)_{d=1,2,\dots,D}$.
We assume that, for each $d \in \{1,2,\dots,D\}$,
$(x^d, y^d, z^d)$ is an independent and identical realization of random variables $X$, $Y$, and $Z$, respectively.
Our predictive optimization is summarized in the three phases below.
\paragraph{Feature selection phase}
Feature selection is widely recognized as an important preprocessing phase in machine learning prediction,
in which a feature selector discards useless or redundant features in order to improve prediction performance~\cite{li2017feature}.
We consider here {\it supervised} and {\it batch} feature selection,
which aims to select subset $Z_{\kappa}$ of feature indices on the basis of given data $\cD$.
Let $MB(Y)$ denote a feature selection algorithm that outputs subset $S$ of $X \cup Y$
which is useful for predicting $Y$:
\begin{align*}
	S = MB(Y).
\end{align*}
The selected external features $Z_{\kappa}$ is then defined as $Z_{\kappa} = S \cap Z$,
where $\kappa \subset [1,K]$ is indices of selected features.
Examples of feature selection algorithms include 
feature selection on the basis of mutual information~\cite{brown2012conditional} 
and that on the basis of sparse regression~\cite{tibshirani1996regression,liu2009multi}.
For a thorough review of existing feature selection algorithms,
see~\cite{li2017feature}.
Here MB is named after Markov blanket,
and our discussion that feature selection algorithms above can be 
regarded as a Markov blanket discovery algorithm
is presented in Section~\ref{subsec_blanket}.

\paragraph{Prediction phase}
Given selected feature indices $\kappa$, a loss function $\ell$, and a hypothesis space $\cF_{\kappa}$,
the prediction phase in general computes a regression function $\hat{f}_{\kappa}: \cX \times \bR^{\kappa} \to \bR^N$ in $\cF_{\kappa}$
that minimizes the empirical loss:
\begin{align*}
\hat{f}_{\kappa} := \argmin_{f_{\kappa} \in \cF_{\kappa}} \sum_{i=1}^{d} \ell \left(y^i, f_{\kappa}(x^i,z_{\kappa}^i) \right).
\end{align*}
In our experiments, we adopt least square loss and linear regression functions.
\paragraph{Optimization phase}
We assume that, before optimization,
specific realization of external features $Z$ is available.
In a price optimization setting, for example,
after such external features as weather, temperature, etc. are revealed,
a storekeeper decides on prices for the day.
The optimization phase thus computes optimized strategies on the basis of 
an estimated prediction formula $\hat{f}_{\kappa}$ and a realization $\tilde{z}$ of $Z$.
Though a simple non-robust formulation can be given by replacing $f$ with $\hat{f}_{\kappa}$ and $\tilde{z}$ with $\tilde{z}_{\kappa}$ in~\eqref{idealOpt},
we present a more general robust optimization formulation:
\begin{align}
\min_{x \in \cX} r(x,y) + \lambda g(x, \tilde{z}_{\kappa})  \quad \text{s.t. } y=\hat{f}_{\kappa}(x,\tilde{z}_{\kappa}). \label{robust}
\end{align}
Here $\lambda$ is a scale of robustness in which $\lambda=0$ corresponds to non-robust optimization, 
and $g$ is referred to as a \emph{robust regularizer}.
Discussion of robust optimization is found in Section~\ref{sec_robust_opt}.

\section{Preliminary}
As noted in our introduction,
simple application of feature selection could cause cofounding bias.
This section introduces the language of causality, 
which enables us to characterize the conditions under which cofounding bias disappears.
For simplicity of presentation, this section assumes the well-known \emph{causal sufficiency assumption}~\cite{zhang2002strong} stating that no unobserved cofounder exists,
but our main discussion in the subsequent section does not rely this.
\subsection{Causal Bayesian network}
We introduce here the notion of a causal Bayesian network, which is a standard tool
for describing causal relationships.
Let $\bV= X \cup Y \cup Z$ be a set of random variables.
A \emph{Bayesian network} for $\bV$
is a pair $(G, p)$, where $G=(\bV, E)$ is a directed acyclic graph
and $p$ is a joint distribution over $\bV$, satisfying the following factorization~\cite{pearl2009causality}:
\begin{align*}
p(\bV) = \prod_{V \in \bV} p(V \mid Pa(V)).
\end{align*}
Here vertices are associated with random variables in $G$,
and for $V \in \bV$, $Pa(V):= \{U \in \bV \mid (U,V) \in E\}$ denotes the set of random variables
that are parents in $G$.
We call the network a \emph{causal Bayesian network} (or causal network) if all edges represent causal effects.
Let $\rdo(X=x)$ (or $\rdo(X)$ shortly) denote \emph{intervention},
which is an operation fixing the realization of random variable $X$ to $x$ regardless of the joint distribution $p$.
In our predictive optimization setting, 
an optimizer intervenes on the decision variables $X$.
Let $v_X$ denote the projection of vector $\bv$ onto coordinates in $X$.
Given an intervention $\rdo(X=x)$,
a post interventional distribution can also be factorized to accord with the network~\cite{pearl2009causality}:
\begin{align}
	p(\bV = \bv \mid \rdo(X=x))  
	=
	\begin{cases}
		\prod_{V \in \bV \setminus X} p(V = v \mid Pa(V),) \quad  \text{ if $v_X=x$},\\
		0 \quad \text{ otherwise.}
	\end{cases}
	\label{factorization}
\end{align}
In general, post interventional conditional distribution $p(Y \mid \rdo(X), Z_{\kappa})$
might not be equivalent to corresponding conditional distribution $p(Y \mid X, Z_{\kappa})$,
and such a gap could cause cofounding bias in predictive optimization.
\subsection{Adjustment criteria}
Given a causal network, we can compute a post interventional distribution on the basis of the factorization formula~\eqref{factorization}.
Specifically, this can characterize the conditions under which cofounding bias disappears,
which conditions are called \emph{adjustment criteria}~\cite{pearl2009causality}.
We here introduce one of the most basic criteria, referred to as a \emph{back-door criterion}.
We call an ordered tuple $T=(T_1,T_2,\dots,T_j)$ of vertices a {\it path} if either $(T_i,T_{i+1}) \in E$ or $(T_{i+1},T_i) \in E$ holds for every $i=1,2,\dots,j-1$.
It is specifically called a \emph{directed path} if $(T_i,T_{i+1}) \in E$ for every $i$.
For a triplet of nodes $(T_{i-1},T_{i}, T_{i+1})$ in a path,
$T_{i}$ is called a collider if $(T_{i-1},T_{i}), (T_{i+1},T_i) \in E$.
A node which is not a collider is called a noncollider.
If there exists a directed path from $U \in \bV$ to $V \in \bV$,
then $U$ is called an ancestor of $V$, and $V$ is called a descendant of $U$.
The following d-separation is a standard notion in the study of causal inference.
\begin{df}[See~\cite{pearl2009causality}]
	A path $T=(T_1,T_2,\dots,T_j)$ is \emph{d-separated} by $S \subseteq V$	
	if one of the following holds:
	(i) there exists a noncollider in $T$ that is in $S$, or
	(ii) there exists a collider in $T$ that is neither in $S$ nor an ancestor of a node in $S$.
\end{df}
The back-door criterion is then introduced using a d-separation.
\begin{df}[The back-door criterion, see~\cite{pearl2009causality}]
	A set of variables $S \subseteq V$ satisfies the {\em back-door} criterion relative to an ordered pair $(X,Y)$ if 
	no nodes in $S$ are descendants of $X$, and every path between $X$ and $Y$ which contains a directed edge into $X$ is d-separated by $S$.
\end{df}
The back-door criterion characterizes the condition 
under which a conditional distribution and a post-interventional distribution coincide,
and, thus, the cofounding bias disappears.
\begin{thm}[The back-door adjustment, see~\cite{pearl2009causality}]\label{thm_backdoor}
	If $S \subseteq V$ satisfies the back-door criterion relative to $(X,Y)$, then we have
	\begin{align}
	p(Y \mid \rdo(X), S) = p(Y \mid X,S). \label{adjustment}
	\end{align}
\end{thm}
A set of nodes $S$ satisfying~\eqref{adjustment} is called an \emph{adjustment set} (relative to $(X,Y)$).
For a more general discussion about adjustment criteria,
see~\cite{maathuis2015generalized,perkovic2015complete}.
\subsection{Direct cause discovery}
An adjustment set can be computed given the structure of a causal graph,
but in practice this structure will be unknown.
Causal discovery algorithms can then help us in estimating it using 
observational data.
In general, causal discovery algorithms estimate the entire structure of a causal graph,
but for our purposes,
we can focus on direct cause discovery,
since direct causes are desirable from both viewpoints of cofounding bias and prediction performance, explained as follows.
In a predictive optimization setting,
the target variable $Y$ is revealed after the realization of $X$ and $Z$,
and this temporal context implies the following restriction on the network structure.
\begin{assump}\label{assumpTemporal}
	No nodes in $X$ and $Z$ are descendants of $Y$ in $G$.
\end{assump}
This assumption, together with the back-door adjustment, implies that
any set $S$ that includes \emph{direct causes} $Pa(Y)$ of $Y$ is an adjustment set.
\begin{fact}\label{fact}
	If Assumption~\ref{assumpTemporal} holds and $S \supseteq Pa(Y)$,
	then $Z_{\kappa} := S \cap Z$ satisfy the back-door criterion relative to $(X,Y)$. In particular, $Z_{\kappa}$ is an adjustment set.
\end{fact}
Such direct causes $Pa(Y)$ are known to be desirable
also in terms of feature selection
for achieving good prediction performance~\cite{aliferis2010local1,koller1996toward},
and among the various adjustment sets,
the set of direct causes is one of the most promising candidates for predictive optimization.

The majority of existing causal discovery algorithms are based on the following faithfulness assumption.
\begin{df}
	A causal network $(G,p)$ is faithful if every conditional independence in $p$ is read from
	a d-separation in $G$, in other words, for any $U,V \in \bV$ and $S \subseteq \bV$, $U \indepe V \mid S$ if and only if every path between $U$ and $V$ is d-separated by $S$.
\end{df}
Given faithfulness,
the direct causes are characterized by the following conditional independence.
\begin{prop}[{\cite{pearl2014probabilistic}, see~\cite{pena2007towards}[Theorem 3] also)}]\label{adjustmentDirectCause}
	If Assumption~\ref{assumpTemporal} and faithfulness holds,
	then a set $S \subseteq \bV \setminus Y$ satisfy $Pa(Y) \subseteq S$ if and only if
	\begin{align}
	Y \indepe \bV \setminus S \mid S. \label{directCause}
	\end{align}
\end{prop}
Given faithfulness, thus,
the direct causes $Pa(Y)$ are characterized as a minimal set satisfying the conditional independence~\eqref{directCause} in our setting.
In general,
such a minimal set satisfying~\eqref{directCause} is called a \emph{Markov blanket},
and we can utilize existing Markov blanket discovery algorithms: examples include~\cite{tsamardinos2003algorithms,pena2007towards,margaritis2000bayesian,tsamardinos2003time}.

One drawback of the approach on the basis of direct cause discovery is that it is essentially dependent on the faithfulness assumption.
\begin{rmk}\label{rmk_counter}
	We here show an example consisting of three variables for which, without faithfulness,
	conditional independence~\eqref{directCause} cannot guarantee the discovery of direct causes.
	Let us consider causal network $(G,p)$ with three variables $\bV = \{X, Y, Z \}$ and
	three edges $E = \{(X,Y), (Z,X), (Z,Y) \}$, as seen in Figure~\ref{fig1}.
	Suppose that it identically holds that $X=Z$.
	It then holds that $Y \indepe Z \mid X$. However, the singleton set $\{X \}$ do not include the direct cause of $Y$.
	Also, in larger networks in practice, a similar setting might occur when a decision variable can be completely explained by external features.
	
	The above example also illustrates that,
	even if the underlying distribution is faithful, if it is \emph{almost unfaithful} then
	a direct cause discovery algorithm might in practice incur difficulty 
	in correctly determining conditional independence.
	Such a practical requisite condition is successfully characterized by the notion of $\lambda$-strong faithfulness~\cite{kalisch2007estimating,uhler2013geometry,zhang2002strong}
	in a normal distribution setting.
\end{rmk}
\subsection{Relationship between Markov blanket discovery and feature selection}\label{subsec_blanket}
In predictive optimization, Assumption~\ref{assumpTemporal} reduces the problem of finding a set of direct causes into that of finding a Markov blanket.
The relationship between Markov blanket discovery and feature selection 
has been studied~\cite{aliferis2010local1,koller1996toward}.
We here briefly review this relationship,
and demonstrate that some of feature selection algorithms can be regarded as
approximate Markov blanket discovery algorithms.

In the context of causal discovery,
Markov blanket discovery algorithms try to compute a set of variables $S$
that achieve the conditional independence $Y \indepe (\bV \setminus S) \mid S$.
In practice, it can find a correct Markov blanket only when a sufficient amount of data
are available so as to correctly compute a series of conditional independence tests.
An example of such an algorithm is IAMB~\cite{tsamardinos2003algorithms},
which shows that a simple greedy forward and backward algorithm can compute
a Markov blanket given sufficient amount of data.

In the context of feature selection,
a sparse feature selection algorithm~\cite{tibshirani1996regression,liu2009multi}
can find a minimal set that is linearly dependent on the target variable,
given a sufficient number of samples.
Mutual-information-based feature selection~\cite{brown2012conditional}
greedily finds a minimal set of variables $S$ that achieve $I(V ,Y \mid S) \approx 0$
for every $V \in \bV \setminus S$, and one variant adopts a forward and backward search.
These algorithms can be regarded as approximate Markov blanket discovery algorithms,
with approximation of statistical dependence by, respectively,
linear dependence and positive mutual information.

With these observations in mind,
we denote an algorithm  (possibly approximate) for finding a Markov blanket of $Y$ by $MB(Y)$,
and such algorithms include the above sparse feature selection and mutual-information-based feature selection.
Note that all these algorithms also fail in direct cause discovery without faithfulness,
as is shown in Remark~\ref{rmk_counter}.

\section{Causally admissible feature selection}\label{sec_qualitative}
This section presents our first contribution: we prove a novel adjustment criterion
and present a meta-algorithm which utilizes an existing feature selection algorithm 
so as to achieve no cofounding bias even under unfaithfulness.
Our approach relies on the following assumption, rather than on faithfulness.
\begin{assump}\label{assump}
	(i) No nodes in $Z$ are descendants of $Z$, and (ii) $Z$ is an adjustment set relative to $(X,Y)$.
\end{assump}
The first half (i) of this assumption is implied by the temporal context of predictive optimization pipeline as similar to Assumption~\ref{assumpTemporal}.
The second half (ii) is implied by Fact~\ref{fact},
and Fact~\ref{fact} holds by the causal sufficiency, which was assumed in the previous section.
We state this property as an assumption
to maintain the validity of our discussion even without the causal sufficiency assumption.
\begin{thm}\label{thm_extended_backdoor}
	If Assumption~\ref{assump} holds and $Z_{\kappa} \subseteq V$ satisfies $X \indepe (Z\setminus Z_{\kappa} ) \mid Z_{\kappa}$,
	then $Z_{\kappa}$ is an adjustment set relative to $(X,Y)$.
\end{thm}
This criterion leads us to propose Algorithm~\ref{algo_FS}.
Recall that a Markov blanket discovery algorithm $MB$
maps a subset of nodes $W \subseteq \bV$ to $S\subseteq \bV \setminus W$ which  (possibly approximately) satisfies the conditional independence 
$W \indepe (\bV \setminus S) \mid S$.
Given such an algorithm,
the proposed algorithm computes $MB(X \cup Y)$,
rather than the $MB(Y)$ of the previous direct cause discovery approach.
Observe that, in contrast to Proposition~\ref{adjustmentDirectCause} 
given for the previous approach,
Theorem~\ref{thm_extended_backdoor} does not require the faithfulness assumption for characterizing an adjustment set by conditional independence.
This enables our algorithm to compute an adjustment set even under unfaithfulness.
Note that, according to Theorem~\ref{thm_extended_backdoor},
$MB(X)$ is a smaller adjustment set than $MB(X \cup Y)$,
but we compute $MB(X \cup Y)$ so as to simultaneously improve prediction performance,
where this replacement is justified by the following implication~\cite{pearl2014probabilistic} (see~\cite{pena2007towards}[Theorem 1]  also).
\begin{align*}
	(X \cup Y) \indepe (Z \setminus Z_{\kappa}) \mid Z_{\kappa} \Rightarrow X \indepe (Z \setminus Z_{\kappa}) \mid Z_{\kappa}.
\end{align*}
\begin{algorithm}[t]
	\caption{Causality Admissible Feature Selection}\label{algo_FS}
	\begin{algorithmic}[1]
		\REQUIRE Markov blanket discovery algorithm $MB$
		\ENSURE Subset $W \subseteq \bV \setminus Y$ of features.
		\STATE Return $MB(X\cup Y)$.
	\end{algorithmic}
\end{algorithm}
We conclude this section with a discussion on Assumption~\ref{assump}
when the causal sufficiency does not hold.
\begin{rmk}\label{rmkValidity}
	If the causal sufficiency holds, then Assumption~\ref{assump} (ii) is implied by
	Fact~\ref{fact}.
	Suppose that the causal sufficiency does not hold, and observable $Z$ is a subset of a entire external feature set $Z^*$ which is causally sufficient.
	According to our proof of Theorem~\ref{thm_extended_backdoor},
	with additional assumption $X \indepe (Z^* \setminus Z) \mid Z$,
	our adjustment criterion is still valid.
	Intuitively speaking,
	this additional condition requires that the features which have influenced
	a human decision-maker giving $X$ in historical data are successfully collected in $Z$,
	which means that these features have been at least noticed by the decision-maker,
	and thus are more plausible than the causal sufficiency assumption.
\end{rmk}

\section{Robust optimization using adjustment sets}~\label{sec_robust_opt}
The previous section proposed the utilization of features $MB(X \cup Y)$
instead $MB(Y)$ for avoiding cofounding bias,
but the features $MB(X \cup Y) \setminus MB(Y)$ are discarded in $MB(Y)$
since they are redundant and useless from the viewpoint of prediction.
This section then presents our second contribution:
revealing the role of the redundant features in a predictive optimization pipeline.
\subsection{Generalized bias-variance decomposition}
This section slightly generalizes a well-known bias-variance decomposition
for the explicit representation of cofounding bias.
Let us define define $\overline{Y}_{X,Z_{\kappa}} := E[Y \mid X,Z_{\kappa}]$ and $\overline{Y}_{\rdo(X),Z_{\kappa}} := E[Y \mid \rdo(X),Z_{\kappa}]$.
We also define the optimal predictor $f^*_{\kappa}$ by
$f^*_{\kappa} := {\argmin}_{f_{\kappa} \in \cF_{\kappa}} E\left[\ell \left(Y, f_{\kappa}(X,Z_{\kappa}) \right)\right]$.
For each $z \in \bR^K$, then, a well-known bias-variance decomposition shows
\begin{align*}
&E_{Y, \cD}[\|Y - \hat{f}_{\kappa}(X,Z_{\kappa}) \|^2 \mid \rdo(X=x),Z]   \\
&= \underbrace{E_{Y}[\|Y - \overline{Y}_{\rdo(x),z}  \|^2 \mid \rdo(X), Z ]}_{\mathrm{noise}} 
 +  \underbrace{\|\overline{Y}_{\rdo(x),z}- f^*_{\kappa}(x,z_{\kappa}) \|^2}_{\mathrm{bias}}  + \underbrace{E_{\cD}[\|f^*_{\kappa}(x,z_{\kappa}) - \hat{f}_{\kappa}(x,z_{\kappa})  \|^2 ]}_{\mathrm{variance}}. 
\end{align*}
Here $E_{\cD}$ is the expectation with respect to the historical data.
Let us consider further decomposition of the bias term into cofounding bias and prediction bias, described as
\begin{align*}
\|\overline{Y}_{\rdo(x),z}- f^*_{\kappa}(x,z_{\kappa}) \|^2 
= \|\underbrace{(\overline{Y}_{\rdo(x),z} - \overline{Y}_{x,z_{\kappa}}) }_{\mathrm{cofounding \  bias}} + \underbrace{(\overline{Y}_{x,z_{\kappa}} - f^*_{\kappa}(x,z_{\kappa}))}_{\mathrm{prediction \ bias}} \|^2 . 
\end{align*}
\subsection{Transforming causality bias into statistical variance using redundant features}
We define the sum $C_{x,z}(\kappa)$ of cofounding bias and variance given $\rdo(X=x)$ and $Z=z$ as:
\begin{align*}
C_{x,z}(\kappa) 
:= \| \overline{Y}_{\rdo(x),z} - \overline{Y}_{x,z_{\kappa}} \|^2 
+ E_{\cD}[\|f^*_{\kappa}(x,z_{\kappa}) - \hat{f}_{\kappa}(x,z_{\kappa})  \|^2 ].
\end{align*}
The following statement reveals that this sum is constant under certain assumptions.
\begin{prop}
	For $\kappa_1$ and $\kappa_2$, assume that 
	(i) $\hat{f}_{\kappa_1}(x,z_{\kappa_1}) = \hat{f}_{\kappa_2}(x,z_{\kappa_2})$ for every $\cD$, $x$, and $z$, and (ii) both $\hat{f}_{\kappa_1}$ and $\hat{f}_{\kappa_2}$ are unbiased estimators (having no prediction bias).
	It holds, then, that $C_{x,z}(\kappa_1) = C_{x,z}(\kappa_2)$.
\end{prop}
Although the above assumptions cannot be precisely satisfied in reality,
the statement offers important qualitative observations.
Assume that $Z_{\kappa_1} = MB(Y) \cap Z$ which follows existing approaches,
and $Z_{\kappa_2} = MB(X \cup Y)$, which follows our proposed approach.
Assumption (i) requires that both $Z_{\kappa_1}$ and $Z_{\kappa_2}$ are sufficient for prediction,
so that the predictors $\hat{f}_{\kappa_1}$ and $\hat{f}_{\kappa_2}$ are the same,
and assumption (ii) requires unbiasedness in the predictor, which is a common assumption in prediction.
Under the assumptions of sufficiency and unbiasedness,
the statement confirms that 
the redundant features are in fact useless in terms of prediction error, which is the sum of bias and variance.
In terms of optimization, however,
variance is far preferable to bias:
variance is measurable in practice by means of such methods as bootstrap sampling~\cite{efron1994introduction},
and risky strategies with high variance can be avoided with the aid of a robust optimization technique.
The useless features can exchange causality bias for variance,
and, thus, are useful in terms of optimization.
\subsection{Avoiding high-variance strategies by means of robust optimization}\label{subsec_robust_opt}
Our meta-algorithm transforms cofounding bias into variance,
and, thus, is effective only when combined with a robust optimization technology that can regularize high variance strategies.
This section briefly introduces existing robust optimization technologies
applicable to predictive optimization.
One of the most standard formulations of robust optimization  is given by defining $g$ in \eqref{robust} as a variance of an objective function: 
\begin{align*}
	g_{\Var}(x) := \Var_{\cD} [r(x,\hat{f}(x,\tilde{z})). 
\end{align*}
For linear programming~\cite{ben1999robust} and certain case in quadratic programming~\cite{yabe2017robust} on the basis of linear regression,
explicit forms of $g$ and efficient optimization algorithms are available.
For a survey of robust optimization technologies, see~\cite{ben2009robust,bertsimas2011theory}.

\begin{figure*}[t]
	\begin{minipage}[t]{.33\textwidth}
		\centering
		\includegraphics[width=\linewidth]{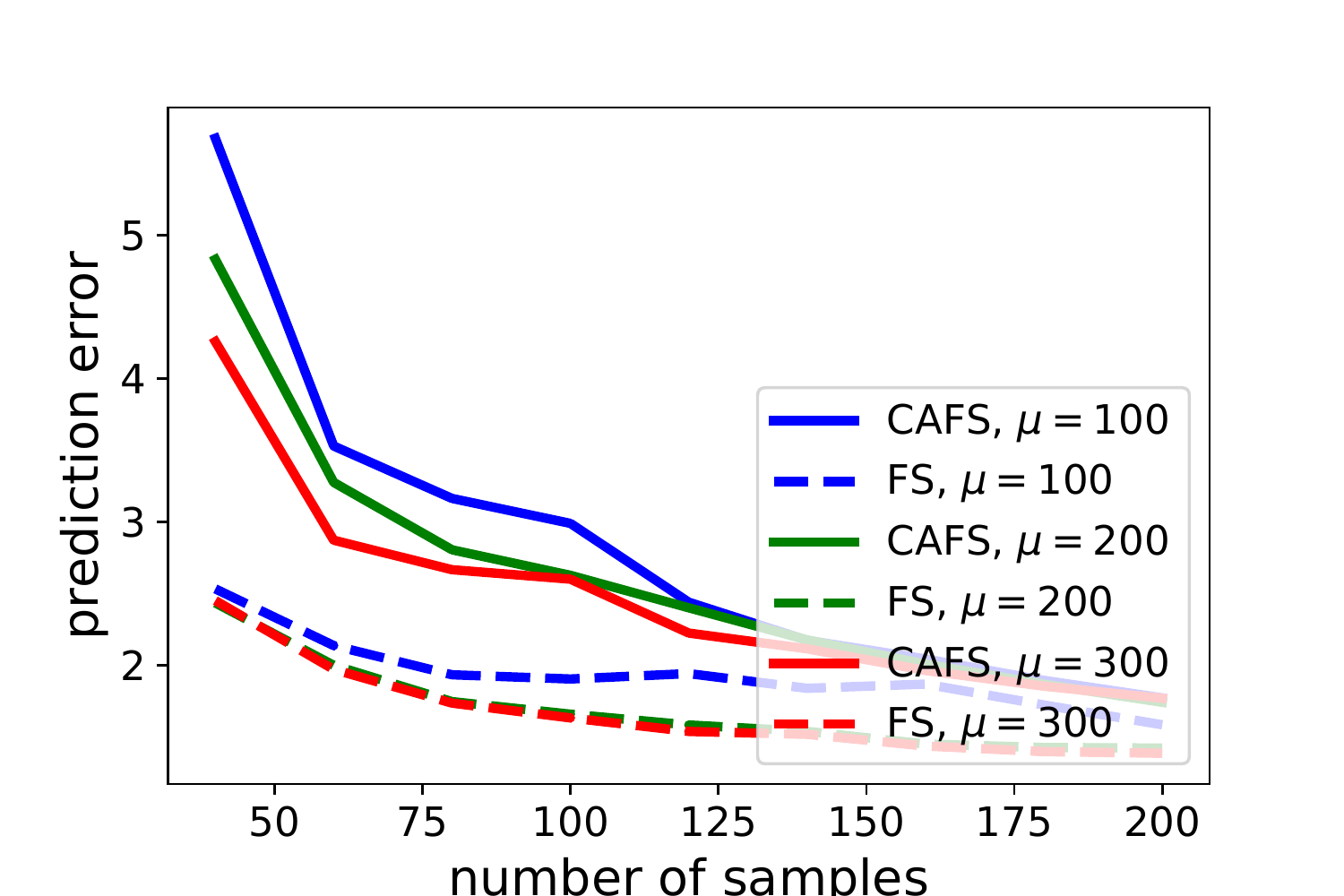}
		\caption{The result of prediction. The horizontal axes show the number of historical samples, and the vertical axes show the prediction error. The solid and dotted lines show the respective results for $\rCF$ and $\rFS$.
			The blue, green, and red lines show the results with scale parameters $\mu=100$, $200$, $300$, respectively.}\label{Exp2}
	\end{minipage}
	\hspace{0.01\hsize}
	\begin{minipage}[t]{.66\textwidth}
		\begin{minipage}[t]{.49\textwidth}
			\includegraphics[width=\linewidth]{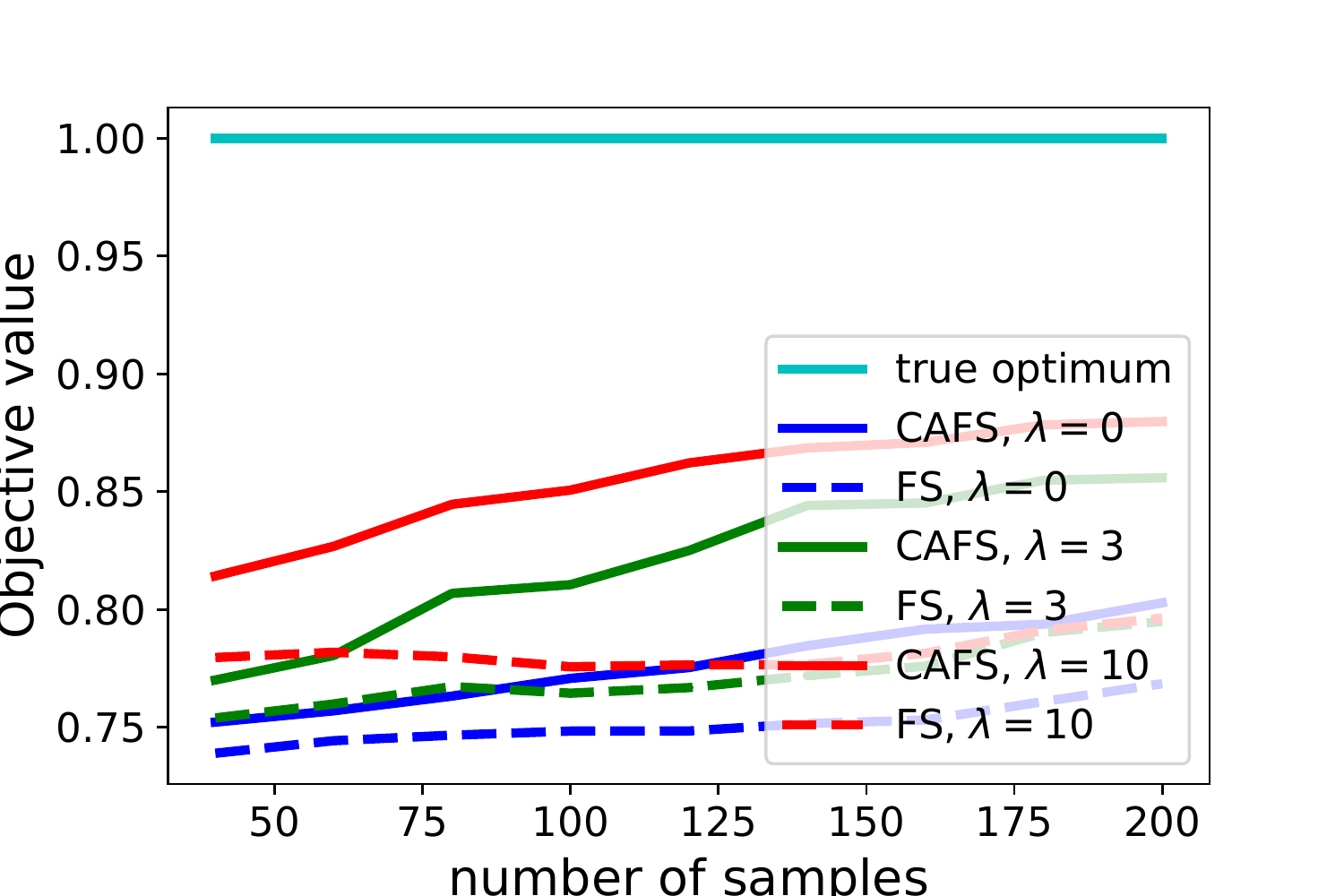}
		\end{minipage}
		\begin{minipage}[t]{.49\textwidth}
			\includegraphics[width=\linewidth]{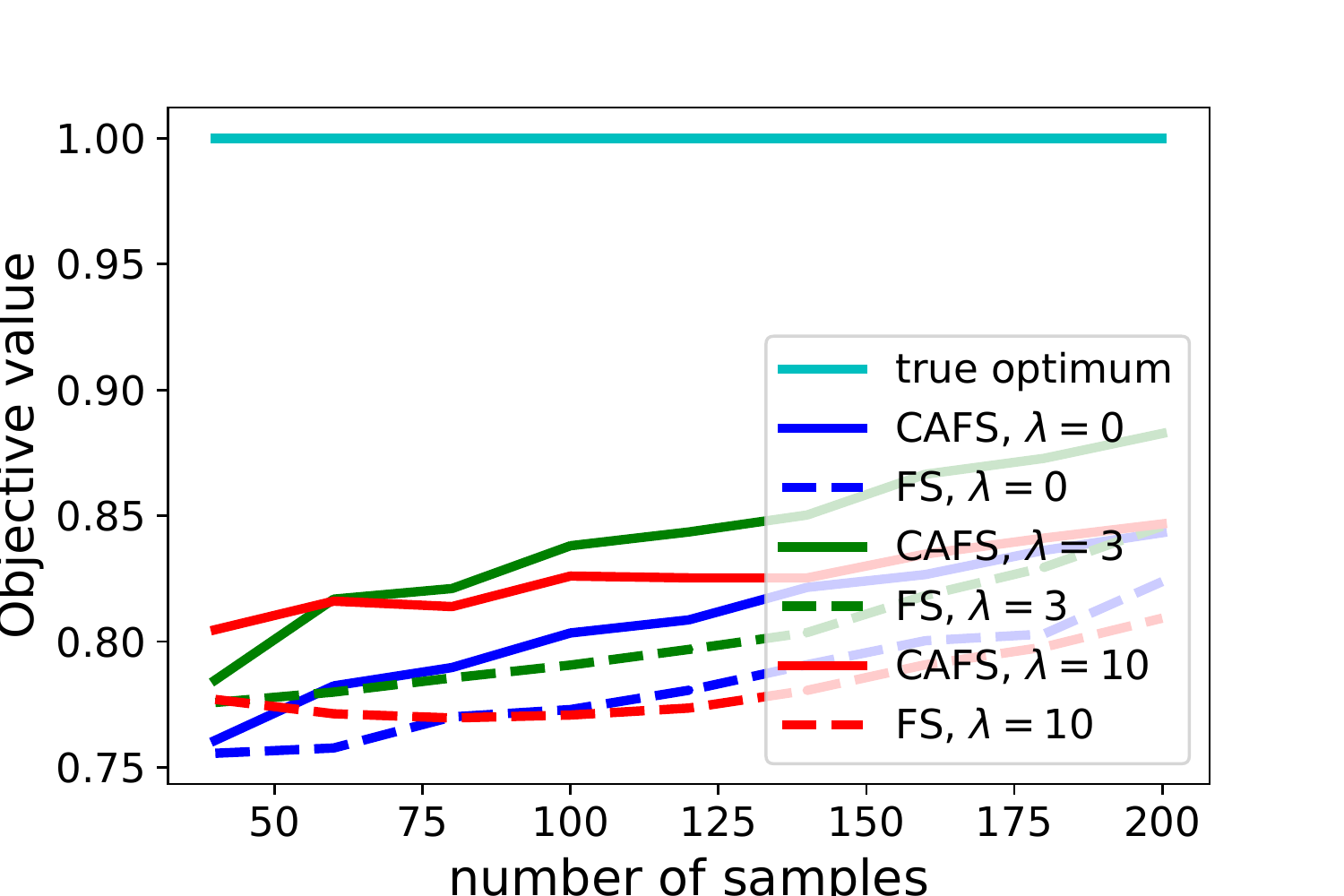}
		\end{minipage}
		\caption{The objective value of robust optimum strategies. The horizontal axis shows the number of historical samples,
			and the vertical axis shows the true objective value. The aquamarine line shows the true optimum value,
			and solid and dotted lines respectively show the values of $\rCF$ and $\rFS$. The blue, green, and red lines respectively show the scale parameters $\lambda = 0$, $3$, and $10$. Left and right figure respectively show the result with parameter $\alpha = 0.1$ and $\alpha =0.5$.}\label{Exp1}
	\end{minipage}
	\vspace{-5truemm}
\end{figure*}

\section{Experiments}\label{sec_experiments}
This section shows the performance of our causally admissible feature selection
framework through experiments in predictive price optimization problem~\cite{yabe2017robust} using synthetic data.
\subsection{Problem setting of price optimization}
Let $M$ be the number of products, and let $x \in \bR^M$ denote a price vector that is a set of decision variables, 
$y \in \bR^N$ (where $N=M$) denote the demand vector that is the target variables,
and $z \in \bR^K$ denote external features (temperature, weather, weekday or not, etc.).
The goal is to maximize the revenue function which is an inner product of price and demand:
$r(x,y) := x^{\top} y$.
We are given a set of historical point of sales data $\cD = \{(x_d, y_d, z_d) \mid d=1,2,\dots,D \}$ of size $D$ that consists of i.i.d. realizations of $(X, Y, Z)$,
which is generated according to an unknown causal network $(G,p)$.
\paragraph{Synthetic generation of causal network}
We assume the temporal context of predictive optimization 
given in Assumption~\ref{assumpTemporal} and Assumption~\ref{assump} (i).
We also assume here that neither $X$, $Y$, nor $Z$ have internal edges.
For each $m,n=1,2,\dots,M$, and $k=1,2,\dots,K$, 
we generate the graph $G$ according to $(X_m, Y_n) \in E$, $\Prob \left( (Z_k, X_m) \in E \right) = 0.1$,
and $\Prob \left( (Z_k, Y_n) \in E \right) = 0.5$.
\paragraph{Generation of a linear SEM}
Given a causal network $G$ generated above,
we define the joint distribution $p$ by linear structural equation modeling (linear SEM)\cite{pearl2009causality}, 
which is one of the most standard model of Bayesian network.
Let $Ber(p)$ denote the Bernoulli distribution with mean $p$,
and $\cN(0, \sigma^2)$ denote the normal distribution with mean $0$ and variance $\sigma^2$. 
For each $k=1,2,\dots,K$ and $m,n=1,2,\dots,M$, we define our linear SEM as:
\begin{align*}
Z_{k} \sim Ber(p_k), \quad X_{m} = 1 - 0.1 \sum_{k: (Z_k, X_m) \in E} Z_{k} - 0.1 \varepsilon_m, 
\end{align*}
\vspace{-4truemm}
\begin{align*}
Y_{n} &= \sum_{m=1}^M a_{n,m} x_m + b_{n} + \sum_{k: (Z_k, Y_n) \in E} c_{n,k} + \delta_n,
\end{align*} 
where $\varepsilon_m \sim Ber(\alpha)$ with $\alpha \in [0,1]$ and $\delta_n \sim \cN(0,100)$.
Parameterization of $p_k$, $A=(a_{n,m})$, $b = (b_n)$, and $C = (c_{n,k})$, and its interpretation are presented in the supplementary material.
\subsection{Algorithms}
We specify the feature selection, prediction, and robust optimization
in the general problem setting in Section~\ref{sec_prob}.

For feature selection phase, we here adopt the sparse feature selection algorithm of~\cite{liu2009multi} and its implementation as presented by the authors of~\cite{li2017feature}.
Given $U \subseteq \bV$ and $V := \bV \setminus U$, 
let $D_{U} \in \bR^{U \times D}$ and $D_{V} \in \bR^{V \times D}$
be respective historical data matrices for variables $U$ and $V$, which is extracted from $\cD$,  
and let $W_V \in \bR^{U \times V}$ be a matrix indexed by $U$ and $V$.
The output of feature selection $MB(U)$ is then defined as the list of nonzero column in the solution $W^*_{V}$ of the following sparse regression:
\begin{align*}
	\min_{ W_{V}} \|D_U - W_V D_V \|_F + \mu \|W_V \|_{1,2},
\end{align*}
where $\mu$ is a scale of the regularizer. 
Selected external features $Z_{\kappa}$ are then defined as $Z_{\kappa} = MB(U) \cap Z$.
We compute $MB$ for the case $U = Y$ and $U= X \cup Y$ in our experiments.

For prediction phase, we adopt the least square estimator. Given selected feature indices $\kappa$,
we estimate the linear prediction model $\hat{y} = \hat{f}_{\kappa}(x, z_{\kappa}) = \hat{A}x + \hat{b} + \hat{C}_{\kappa}z_{\kappa}$ by the least square method.

For optimization phase, we apply the robust optimization technique of~\cite{yabe2017robust} for defining $g$ in \eqref{robust}.
They defined $g(x,z_{\kappa})$ as a variance of objective function,
and proved the explicit form $g(x,\tilde{z}_{\kappa}) := \|\Sigma x \|^{1/2} \|\Sigma' v(x,\tilde{z}_{\kappa}) \|^{1/2}$.
Here $\Sigma$ is the covariance matrix of $Y$, 
and $\Sigma'$ is essentially an inverse of $D'_{X \cup Z_{\kappa}} D_{X \cup Z_{\kappa}}^{'\top}$,
and $v(x,\tilde{z}_{\kappa}) = (x^{\top}, \tilde{z}_{\kappa}, 1)^{\top}$.
Note that, while the original formulation does not deal with external features,
this extension can be directly obtained by first regarding $z_{\kappa}$ also as decision variables and then fixing $z_{\kappa} = \tilde{z}_{\kappa}$.
\subsection{Experimental results}
For each setting, we conducted 50 randomized experiments and took an average over them. 
We fixed the size of the problem as $M = K = 10$. 
\paragraph{Comparison of prediction accuracy}
We first compared the original feature selection $MB(Y)$ (denoted by $\rFS$) and our causally admissible feature selection $MB(X\cup Y)$  (denoted by $\rCF$) in terms of prediction accuracy in Figure~\ref{Exp1}.
We observed that The prediction accuracy of $\rFS$ is better than that of $\rCF$ with every choice of regularization parameter $\mu$, and the gap is huge when the number of available sample is small.
This indicates that the redundant features of $\rCF$ are in fact useless in terms of prediction accuracy, and our modification would not improve, or might even degrade, prediction accuracy.

\paragraph{Efficiency of robust optimum strategy}
We fixed $\mu = 200$ on the basis of the previous experiment,
and we then compared $\rFS$ and $\rCF$ in terms of optimization.
After computing a prediction formula, 
we generated $\tilde{z}$ 10 times and conducted robust optimization for computing an optimized strategy $\tilde{x}$ with several scale $\lambda = 0,3,10$ of robust regularizer in \eqref{robust}.
Here, $\lambda =0$ corresponds to non-robust optimization, and  $\lambda = 10$ computes most conservative pricing strategy.
We computed the true objective value of optimized strategy $\tilde{x}$,
and Figure~\ref{Exp2} shows the average of the performance normalized by the true optimum value, with parameters $\alpha=0.1$ (left) and $\alpha=0.5$ (right).
We observed that:
\begin{itemize}[noitemsep,nolistsep,leftmargin=*]
	\item For $\alpha = 0.1$, $\rCF$ without robust formulation ($\lambda = 0$) is only as good as $\rFS$ with the best parameterization ($\lambda=10$). With robust formulation ($\lambda >0$), however, the performance of $\rCF$ drastically improves, while the performance of $\rFS$ rarely improves.
	This demonstrate that redundant features enable a robust optimizer to distinguish stable strategies from risky ones with high variance.
	\item For $\alpha = 0.5$, although $\rCF$ outperforms $\rFS$, the performance gap is not as huge as with that $\alpha = 0.1$. In particular, with $\lambda=0,3$, the performance of $\rFS$ steadily improves as the number of sample increase, 
	in contrast to little improvement in $\alpha = 0.1$.
	Recall that $\alpha$ controls the Bernoulli independent random noise on historical pricing strategies.
	With $\alpha = 0$, the causal network is unfaithful.
	With $\alpha = 0.5$ which makes the noise have largest entropy, the network is \emph{the most faithful}.
	With this setting, $\rFS$ relying on the faithfulness assumption is less affected by the cofounding bias,
	and thus the performance gap between $\rFS$ and $\rCF$ is not large.
	\item For $\alpha=0.1$, the conservative parameterization $\lambda = 10$ outperforms $\lambda = 0,3$, while mild robustness $\lambda=3$ is basically the best in $\alpha = 0.5$.
	Estimation of regression parameter is much more difficult in $\alpha = 0.1$ because of small independent noise,
	and in such scenario~\cite{yabe2017robust} have been observed that conservative parameterization is preferable.
	In fact, even with $\alpha=0.5$, $\lambda=10$ outperforms $\lambda=3$ with smallest number of samples ($D=40$).
\end{itemize}
Thus our modified feature selection algorithm, together with robust optimization technology,
achieves efficient predictive optimization.

\section{Summary and Future Work}
This paper has proposed a meta-algorithm for use with causally admissible feature selection that can avoid cofounding bias in predictive optimization. 
Our algorithm is based on the contextual restriction of causal network structure
and the novel adjustment criteria that maintain its effectiveness even under a causally unfaithful condition.
Features useless in terms of prediction turn out to be useful in optimization, transforming intractable causality bias into rather tractable prediction variance.
A variance-based regularization technique in robust optimization can then provide safe and effective strategies, as is shown in our experiments.
Future research directions include studies aimed at reducing cofounding bias in other optimization scenarios, also without relying on the faithfulness assumption.

\bibliography{causality}
\bibliographystyle{plain}

\appendix
\section{Proofs}

\begin{proof}[Proof of Fact 5]
	Since every path from $X$ to $Y$ which containes a directed edge into $X$
	must includes a node $P \in Pa(Y) \cap Z \subseteq Z_{\kappa}$ and such $P$
	is noncollider, $Z_{\kappa}$ satisfies the back-door criterion relative to $(X,Y)$.
	Thus $Z_{\kappa}$ is an adjustment set.
\end{proof}

\begin{proof}[Proof of Proposition 7]
	Since $Y$ have no children in the network $G$ by Assumption 4,
	the statement directly follows from Thorem 3 of~\cite{pena2007towards}. 
\end{proof}

\begin{proof}[Proof of Theorem 10]
	Let $U := Z \setminus Z_{\kappa}$. We have 
	\begin{align*}
	p(Y \mid \rdo(X), Z_{\kappa}) 
	= \sum_{U} p(Y \mid , \rdo(X), Z_{\kappa}, U ) p(U \mid \rdo(X), Z_{\kappa}).
	\end{align*}
	Since $Z$ is an adjustment set relative to $(X,Y)$, we have
	\begin{align*}
	p(Y \mid \rdo(X), Z, U) = P(Y \mid X, Z, U).
	\end{align*}
	Since no node in $B$ is a descendant of $X$ 
	and	since $Z_{\kappa}, U \subseteq Z$, we have
	\begin{align*}
	p(U \mid \rdo(X), Z_{\kappa})= P(U \mid Z_{\kappa}).
	\end{align*}
	Further, by the conditional independence, we have $P(U \mid Z_{\kappa}) = P(U \mid  X, Z_{\kappa})$.
	Thus, we have
	\begin{align*}
	P(Y \mid \rdo(X),  Z) &= \sum_{U} P(Y \mid X, Z_{\kappa}, U ) P(U \mid X, Z_{\kappa})\\
	&= P(Y \mid X, Z_{\kappa}).
	\end{align*}
\end{proof}

\begin{proof}[Proof of Proposition 12]
	By the assumption (i), it holds that 
	\begin{align*}
	E_{Y,\cD} [Y - \hat{f}_{\kappa_1}(X, Z_{\kappa_1}) \mid \rdo(X),Z] 
	= E_{Y,\cD} [Y - \hat{f}_{\kappa_2}(X, Z_{\kappa_2}) \mid \rdo(X),Z].
	\end{align*}
	Since by assumption (ii), there exists no the prediction bias, it holds that
	\begin{align*}
	E_{Y,\cD} [Y - \hat{f}_{\kappa_i}(X, Z_{\kappa_i}) \mid \rdo(X),Z]
	= E_{Y}[\|Y_{\rdo(x),z} - \overline{Y}_{\rdo(x),z} \|^2] + C_{x,z}(\kappa_i)
	\end{align*}
	for $i=1,2$.
	These equalities imply the statement.
\end{proof}

\section{Parameterization of SEM in experiments}
Let $\cU$ denote the uniform distribution over $[0,1]$, $Ber(p)$ denote the Bernoulli distribution with mean $p$,
and $\cN(0, \sigma^2)$ denote the normal distribution with mean $0$ and variance $\sigma^2$. 
For each $k=1,2,\dots,K$, $m=1,2,\dots,M$, and $n=1,2,\dots,N$, we generate parameters by $p_k \sim \cU$,
and then define a linear SEM as:
\begin{align*}
Z_{k} &\sim Ber(p_k),\\
X_{m} &= 1 - 0.1 \sum_{k: (Z_k, X_m) \in E} Z_{k} - 0.1 \varepsilon_m, \\
Y_{n} &= \sum_{m=1}^M a_{n,m} x_m + b_{n} + \sum_{k: (Z_k, Y_n) \in E} c_{n,k} + \delta_n,
\end{align*} 
where $\varepsilon_m \sim Ber(\alpha)$ and $\delta_n \sim \cN(100)$.
Intuitively speaking, the list price of $X_{m}$ is $1$,
and a storekeeper decides discounting strategy according to the realization of relevant features $Z_k$ satisfying $(Z_k, X_m) \in E$.
The parameter $\alpha$ controls the amount of independent noise on each products, and thus controls the \emph{amount of faithfulness} in this model,
which is discussed in our experimental result.
The coefficient matrix $A=(a_{n,m})$ and constant vector $b = (b_n)$ are generated in the same way as the experiments of~\cite{yabe2017robust},
and each element in $C = (c_{n,k})$ is generated with the same distribution as that of nondiagonal elements of $A$.

\end{document}